\documentclass[10pt, a4paper]{article}

\usepackage[final]{lrec2026} % this is the new style
\usepackage[T1]{fontenc}
\usepackage[utf8]{inputenc}

\usepackage{microtype}

\usepackage{graphicx}

\usepackage{amssymb}

\usepackage[pro]{fontawesome5}

\usepackage{enumitem}
\usepackage{listings}
\usepackage{booktabs} % For professional table formatting

\usepackage{float}      % In your preamble
\usepackage{placeins}   % In your preamble (if needed)

\usepackage{array}

\usepackage{xcolor}
\usepackage{hyperref}
\usepackage{xurl}

\usepackage{tikz}
\usepackage{pgf-pie}

\lstdefinestyle{mypromptstyle}{
    basicstyle=\ttfamily\fontsize{8.5pt}{10pt}\selectfont,      % Monospaced font in small size
    escapeinside={(*@}{@*)},
    backgroundcolor=\color{gray!10},   % Light gray background
    frame=single,                     % Single-line border around the prompt
    breaklines=true,                  % Wrap long lines
    keepspaces=false,                  % Preserve all whitespace as typed
    columns=fullflexible,                     % Ensures each character occupies the same width
    belowskip=0pt,
    breakindent=0pt,
    moredelim=**[is][\ttfamily\bfseries]{<|}{|>},
    literate={<}{{\texttt{<}}}1 {>}{{\texttt{>}}}1
}

\title{Bridging the Domain Divide: Supervised vs. Zero-Shot Clinical Section Segmentation from MIMIC-III to Obstetrics}

\name{Baris Karacan, Barbara Di Eugenio, Patrick Thornton}

\address{University of Illinois Chicago\\
Chicago, IL, USA\\
bkarac3@uic.edu, bdieugen@uic.edu, patrickcnm@gmail.com}

\abstract{
Clinical free-text notes contain vital patient information. % organized into sections. 
They are structured into labelled sections; 
recognizing these sections has been shown to support clinical decision-making and downstream NLP tasks.  
In this paper, we advance clinical section segmentation through three key contributions.
First, we curate a new de-identified, section-labeled  obstetrics notes dataset, to supplement the medical domains covered in public corpora such as MIMIC-III, on which 
most existing segmentation approaches % rely on supervised models trained 
are trained.
Second, we systematically evaluate transformer-based supervised models for section segmentation on %MedSecId, 
a curated subset of MIMIC-III (in-domain), and on the new obstetrics dataset (out-of-domain). 
Third, we conduct the first head-to-head comparison of supervised models for medical section segmentation with zero-shot large language models.
Our results show that while supervised models perform strongly in-domain, their performance drops substantially out-of-domain. In contrast, zero-shot models demonstrate robust out-of-domain adaptability once hallucinated section headers are corrected. These findings underscore the importance of developing domain-specific clinical resources and highlight zero-shot segmentation as a promising direction for applying healthcare NLP beyond well-studied corpora, as long as hallucinations are appropriately managed.
 \\ \newline \Keywords{Healthcare NLP, Section Segmentation, Zero-Shot Learning} }

\begin{document}

\maketitleabstract

\section{Introduction}
Electronic Health Records (EHRs) are widely used in modern healthcare to provide detailed records of patient encounters and their interactions within the healthcare system \citep{holmes2021electronic}. EHR data often contain free-text clinical notes, which are typically organized into sections such as "\textit{Chief Complaint}" and "\textit{History of Present Illness}." Accurately identifying these sections is crucial for downstream natural language processing (NLP) tasks, including entity extraction, information retrieval, and word sense disambiguation \citep{denny2008development}. However, clinical documentation is variable and often lacks standardized formatting. For example, section headers may use a variety of abbreviations (e.g., “\textit{Social History}” versus “\textit{Social Hx}”) or contain typographical errors (e.g., “\textit{Chief Complain}” versus “\textit{Chief Complaint}”) \citep{ganesan2014general}. Such inconsistencies complicate rule-based solutions and motivate more robust machine-learning and deep-learning approaches.

Although supervised models trained on large public corpora (e.g., MIMIC-III \citep{johnson2016mimic}) have shown promise for section segmentation, their domain adaptation capabilities remain uncertain. In particular, obstetrics represents a specialized clinical subdomain with unique documentation styles and limited annotated data. This variability is evident in how physician counseling information is structured, where the same content may be documented under diverse section headers such as "\textit{Impression and Plan}," "\textit{Assessment and Plan}," "\textit{Assessment}," or even abbreviated formats like "\textit{A/P}," "\textit{A\&P}," and "\textit{A: P:}"; we present an example of such a note in Figure~\ref{fig:quote}.

\begin{figure}[t]
\centering
\footnotesize
%\scriptsize
\begin{minipage}{0.97\linewidth}
\begin{tabular}{|p{0.97\linewidth}|}
\hline
\textbf{a/p:} \\
"This is 28 y/o G6P4013 presents for repeat cesarean delivery and bilateral tuba ligation. admit to labor and delivery cbc, type and screen iv fluids consented for repeat cesarean delivery, possible hysterectomy, possible blood transfusion, and bilateral tubal ligation. Hgb 11.1  patient alst ate at 2100on \textit{<DATE>} discussed with the team \textit{<NAME>} MD Obstetrics and Gynecology Resident  Attending note  I saw patient prior to scheduled repeat c/section this morning, reviewed cerner and counseled patient r/b/a. Patient with history of prior c/s x 1 here for RCD and BTL. Labs this morning includes Hgb 11.1, plat 207, bl gp O+ve. Singlton cephalic, posterior placenta. Writtent consent obtained, will proceed when OR is ready. Patient voiced understanding of the plan. \textit{<NAME>}, MD" \\
\hline
\end{tabular}
\caption{Assessment and Plan section from a sample obstetrics note (includes typographical errors and masked identifier tokens).}
\label{fig:quote}
\end{minipage}
\end{figure}

Such variability often introduces out-of-vocabulary or unseen section headers that can significantly hinder deep learning approaches trained on more standardized data. Recent research suggests that LLMs can handle out-of-domain tasks through zero-shot or few-shot prompting \citep{kojima2022large}, but it is unclear how they fare against traditional supervised methods in specialized domains.
To address these gaps, we contribute the following:

\begin{enumerate}
    \item \textbf{Obstetrics Notes Collection (ONC): } We introduce the \textit{Obstetrics Notes Collection (ONC)}, a de-identified dataset of 100 \textit{History \& Physical (H\&P)} obstetrics notes, annotated in collaboration with a domain expert\footnote{The domain expert is the third author of this paper.}. ONC serves as a realistic benchmark for studying section segmentation in underexplored clinical subdomains and is intended for community reuse.
    \item \textbf{Domain-Specific Evaluation of Supervised Models: }We assess whether transformer-based supervised models originally trained on public datasets can effectively generalize to obstetrics notes. By comparing them on in-domain (MedSecId \citep{landes2022new}) and out-of-domain (ONC) data, we highlight the difficulties in transferring knowledge across clinical sub-specialties.
    \item \textbf{Systematic Comparison With Zero-Shot LLMs: }We present the first head-to-head comparison of supervised transformer models and zero-shot LLMs (i.e., Llama, Mistral and Qwen) for clinical section segmentation. Our experiments reveal challenges (e.g., hallucinated section headers) as well as the potential benefits of zero-shot strategies, especially when annotated data are scarce.
\end{enumerate}

The paper is organized as follows.  We discuss related work in Sec.~\ref{sec:related}, our datasets in Sec.~\ref{sec:data}, and our proposed approaches in Sec.~\ref{sec:methods}. We present our experimental results in Sec.~\ref{sec:experiments}, and  conclude with future directions (Sec.~\ref{sec:conclusion}) and  limitations (Sec.~\ref{sec:limit}).

\section{Related Work}
\label{sec:related}
Before the emergence of advanced machine learning and NLP techniques, early approaches to clinical section segmentation primarily relied on rule-based methods. \citet{denny2008development}, for instance, extracted candidate section header strings from a large corpus of "history and physical" (H\&P) notes through pattern-based matching (e.g., detecting strings that end with punctuation or follow certain capitalization patterns). These candidates were then refined in collaboration with clinicians to build a terminology of section headers. However, purely rule-based methods tend to be inflexible and often fail to handle unexpected variations in unstructured, non-standardized text, which constitutes approximately 80\% of the content in EHRs \citep{kong2019managing}.

To overcome the limitations of rule-based approaches, researchers proposed machine learning-based solutions for section segmentation, often framing it as a sequence-labeling task. \citet{li2010section} trained a Hidden Markov Model (HMM) on a clinical corpus to segment 15 predefined section types. \citet{ganesan2014general} employed an L1-regularized multi-class Logistic Regression model to classify each line of a clinical note into one of five roles (start header, continue header, start section, continue section, or footer) and then used the Viterbi algorithm \citep{viterbi1967error} to determine the most probable sequence of labels.

More recent work has been grounded in transformer-based architectures. \citet{zhang2022section} presented a multi-task transformer model that simultaneously identifies section boundaries and assigns medically relevant labels. \citet{saleh2024tocbert} leveraged BioClinicalBERT embeddings \citep{alsentzer2019publicly} and framed section title and subtitle detection as a named entity recognition (NER) task. While not fully transformer-based, \citet{landes2022new} incorporated BERT embeddings as sentence-level representations, which were then processed using a BiLSTM model for sequence modeling and further refined with a Conditional Random Field (CRF) layer to enforce structured predictions across section boundaries.

Most of this research relies on large publicly available datasets such as MIMIC-III. Since producing high-quality annotated data is very resource-intensive, recent work has explored large language models (LLMs) for clinical section segmentation in zero-shot settings. \citet{zhou2024generalizable} evaluated several LLMs, both zero-shot and fine-tuned, across multiple corpora to assess their section-segmentation effectiveness; however, these LLMs were still tested on common public datasets (e.g., MIMIC-III and i2b2 \citep{uzuner2011i2b2}) rather than more specialized clinical domains. 

Hence, it remains unclear how well these approaches generalize to specialized and underutilized domains like obstetrics. Moreover, existing comparative studies often evaluate supervised methods or LLM-based methods solely against each other, leaving a gap in cross-method comparisons in specialized settings. We address this gap by introducing an informative dataset of obstetrics-related H\&P narratives. We propose both supervised and zero-shot approaches for clinical section segmentation, and then evaluate their performance against each other on our newly collected dataset as well as on publicly available annotated corpora. This sheds new light on how different models perform in a specialized medical domain.

\section{Data}
\label{sec:data}
We use \citet{landes2022new}'s publicly available MedSecId corpus to train and evaluate our models. MedSecId comprises 2,002 fully annotated clinical notes from MIMIC-III, specifically designed for clinical section segmentation. Additionally, we introduce ONC, a novel, de-identified dataset of 100 H\&P notes from 50 vaginal birth after cesarean (VBAC) and 50 repeat cesarean section (RCS) patients to evaluate model performance in obstetrics, an underrepresented clinical domain. All clinical notes in both datasets are written in English.

\subsection{MedSecId}
MedSecId spans five note types: \textit{Discharge summary (1,254)}, \textit{Physician (288)}, \textit{Radiology (205)}, \textit{Echo (198)} and \textit{Consult (57)}. It segments each note into 50 section categories, plus a "\textit{<none>}" label for text outside any predefined section. These categories were established through an iterative physician-led annotation process that combined rule-based pre-labeling with manual revisions \citep{landes2022new}. Using the existing dataset, we split each note into its corresponding sections based on section spans provided by MedSecId. Next, we tokenized each section into lists of sentences using the NLTK sentence tokenizer \citep{bird2009natural}, ensuring each sentence was correctly assigned to its respective section and appeared in the correct order.

\subsection{Obstetrics Notes Collection (ONC)}
ONC was gathered and managed using REDCap \citep{harris2009research, harris2019redcap}. Since the notes contained protected health information (PHI), we transferred them to a HIPAA\footnote{ HIPAA is a U.S. federal law that sets national standards for protecting sensitive patient health information.}-secure environment and automatically de-identified them using the Spark NLP framework \citep{kocaman2021spark}. This framework masked entities, including \textit{NAME}, \textit{LOCATION (address, city, zip code)}, \textit{DATE}, \textit{CONTACT (phone numbers, email addresses)}, and \textit{ID (social security number, medical record number)}. We then manually reviewed all notes to confirm complete PHI removal, and the de-identification process was subsequently validated by the institution's HIPAA privacy office to ensure full compliance.\footnote{ The de-identified ONC dataset and corresponding section annotations are available at: \url{https://github.com/barikosan/obstetric-notes-collection/}.}

Due to annotation resource constraints, we focused on 100 high-quality, full-length H\&P notes from distinct patients across both delivery groups (VBAC and RCS). Annotations involved verifying and labeling section boundaries and header types in collaboration with the same midwifery domain expert. Each section was then split into sentences following the same procedure used for MedSecId. Unlike MedSecId, which is used for both training and evaluation, we employed ONC solely for evaluation and cross-domain benchmarking due to its limited size. While MedSecId includes a mix of general-purpose clinical note types, ONC is obstetrics-specific and incorporates domain-relevant section headers (e.g., "\textit{Pregnancy History}," or "\textit{Gynecologic History}") that capture obstetrics-specific content such as gravida/para notation\footnote{ \textit{Gravida}: total pregnancies and \textit{para}: the number of births reaching viability.} and neonatal outcomes. Rather than normalizing to MedSecId's schema, we retained these headers to preserve the narrative structure and semantics of obstetrics H\&P narratives. 

% \subsection{Additional Details}
To enable fair cross-domain evaluation, we excluded obstetrics-specific headers when testing supervised models trained on MedSecId. This allows us to isolate the models' ability to generalize to a clinically distinct domain. Table~\ref{tab:table_1} compares section headers across both datasets, highlighting shared and domain-specific labels. Across both corpora, we identify a total of 64 distinct section headers, of which 34 appear only in MedSecId, 13 appear only in ONC, and 17 are shared between the two datasets. Table~\ref{tab:table_2} presents the frequency distribution of section spans observed in the ONC corpus.

\begin{table}[t]
    \centering
    \footnotesize
    \resizebox{\columnwidth}{!}{
    \begin{tabular}{lcc}
        \hline
        \textbf{Section Header} & \textbf{MedSecId} & \textbf{ONC Dataset} \\
        \hline
        <none>                         & \textcolor{green}{\faCheckCircle} & \textcolor{green}{\faCheckCircle} \\
        24-hour-events               & \textcolor{green}{\faCheckCircle} & \textcolor{darkgray}{\faTimesCircle} \\
        addendum                      & \textcolor{green}{\faCheckCircle} & \textcolor{darkgray}{\faTimesCircle} \\
        allergies                     & \textcolor{green}{\faCheckCircle} & \textcolor{green}{\faCheckCircle} \\
        assessment-and-plan           & \textcolor{green}{\faCheckCircle} & \textcolor{green}{\faCheckCircle} \\
        chief-complaint               & \textcolor{green}{\faCheckCircle} & \textcolor{green}{\faCheckCircle} \\
        clinical-implications         & \textcolor{green}{\faCheckCircle} & \textcolor{darkgray}{\faTimesCircle} \\
        consent         & \textcolor{darkgray}{\faTimesCircle} & \textcolor{green}{\faCheckCircle} \\
        code-status                   & \textcolor{green}{\faCheckCircle} & \textcolor{darkgray}{\faTimesCircle} \\
        communication                 & \textcolor{green}{\faCheckCircle} & \textcolor{darkgray}{\faTimesCircle} \\
        comparison                    & \textcolor{green}{\faCheckCircle} & \textcolor{darkgray}{\faTimesCircle} \\
        conclusions                   & \textcolor{green}{\faCheckCircle} & \textcolor{darkgray}{\faTimesCircle} \\
        contrast                      & \textcolor{green}{\faCheckCircle} & \textcolor{darkgray}{\faTimesCircle} \\
        critical-care-attending-addendum & \textcolor{green}{\faCheckCircle} & \textcolor{green}{\faCheckCircle} \\
        current-medications           & \textcolor{green}{\faCheckCircle} & \textcolor{green}{\faCheckCircle} \\
        discharge-condition           & \textcolor{green}{\faCheckCircle} & \textcolor{darkgray}{\faTimesCircle} \\
        discharge-diagnosis           & \textcolor{green}{\faCheckCircle} & \textcolor{darkgray}{\faTimesCircle} \\
        discharge-disposition         & \textcolor{green}{\faCheckCircle} & \textcolor{darkgray}{\faTimesCircle} \\
        discharge-instructions        & \textcolor{green}{\faCheckCircle} & \textcolor{darkgray}{\faTimesCircle} \\
        discharge-medications         & \textcolor{green}{\faCheckCircle} & \textcolor{darkgray}{\faTimesCircle} \\
        disposition                   & \textcolor{green}{\faCheckCircle} & \textcolor{darkgray}{\faTimesCircle} \\
        facility                      & \textcolor{green}{\faCheckCircle} & \textcolor{darkgray}{\faTimesCircle} \\
        family-history                & \textcolor{green}{\faCheckCircle} & \textcolor{green}{\faCheckCircle} \\
        findings                      & \textcolor{green}{\faCheckCircle} & \textcolor{darkgray}{\faTimesCircle} \\
        flowsheet-data-vitals         & \textcolor{green}{\faCheckCircle} & \textcolor{darkgray}{\faTimesCircle} \\
        gestational-age         & \textcolor{darkgray}{\faTimesCircle} & \textcolor{green}{\faCheckCircle} \\
        gynecological-history         & \textcolor{darkgray}{\faTimesCircle} & \textcolor{green}{\faCheckCircle} \\
        history                       & \textcolor{green}{\faCheckCircle} & \textcolor{darkgray}{\faTimesCircle} \\
        history-of-present-illness    & \textcolor{green}{\faCheckCircle} & \textcolor{green}{\faCheckCircle} \\
        history-of-present-pregnancy    & \textcolor{darkgray}{\faTimesCircle} & \textcolor{green}{\faCheckCircle} \\
        hospital-course               & \textcolor{green}{\faCheckCircle} & \textcolor{darkgray}{\faTimesCircle} \\
        image-type                    & \textcolor{green}{\faCheckCircle} & \textcolor{darkgray}{\faTimesCircle} \\
        imaging                       & \textcolor{green}{\faCheckCircle} & \textcolor{green}{\faCheckCircle} \\
        impression                    & \textcolor{green}{\faCheckCircle} & \textcolor{darkgray}{\faTimesCircle} \\
        impression-and-plan                    & \textcolor{darkgray}{\faTimesCircle} & \textcolor{green}{\faCheckCircle} \\
        indication                    & \textcolor{green}{\faCheckCircle} & \textcolor{darkgray}{\faTimesCircle} \\
        infusions                     & \textcolor{green}{\faCheckCircle} & \textcolor{darkgray}{\faTimesCircle} \\
        labs                          & \textcolor{green}{\faCheckCircle} & \textcolor{green}{\faCheckCircle} \\
        labs-imaging                  & \textcolor{green}{\faCheckCircle} & \textcolor{green}{\faCheckCircle} \\
        last-dose-of-antibiotics      & \textcolor{green}{\faCheckCircle} & \textcolor{darkgray}{\faTimesCircle} \\
        major-surgical-or-invasive-procedure & \textcolor{green}{\faCheckCircle} & \textcolor{darkgray}{\faTimesCircle} \\
        medical-condition             & \textcolor{green}{\faCheckCircle} & \textcolor{darkgray}{\faTimesCircle} \\
        medication-history            & \textcolor{green}{\faCheckCircle} & \textcolor{darkgray}{\faTimesCircle} \\
        obstetrical-and-gynecological-history            & \textcolor{darkgray}{\faTimesCircle} & \textcolor{green}{\faCheckCircle} \\
        obstetrical-history            & \textcolor{darkgray}{\faTimesCircle} & \textcolor{green}{\faCheckCircle} \\
        other-medications             & \textcolor{green}{\faCheckCircle} & \textcolor{darkgray}{\faTimesCircle} \\
        past-medical-history          & \textcolor{green}{\faCheckCircle} & \textcolor{green}{\faCheckCircle} \\
        past-surgical-history         & \textcolor{green}{\faCheckCircle} & \textcolor{green}{\faCheckCircle} \\
        patient-test-information      & \textcolor{green}{\faCheckCircle} & \textcolor{darkgray}{\faTimesCircle} \\
        physical-examination          & \textcolor{green}{\faCheckCircle} & \textcolor{green}{\faCheckCircle} \\
        plan          & \textcolor{darkgray}{\faTimesCircle} & \textcolor{green}{\faCheckCircle} \\
        pregnancy-history          & \textcolor{darkgray}{\faTimesCircle} & \textcolor{green}{\faCheckCircle} \\
        prenatal-care              & \textcolor{darkgray}{\faTimesCircle} & \textcolor{green}{\faCheckCircle} \\
        prenatal-history              & \textcolor{darkgray}{\faTimesCircle} & \textcolor{green}{\faCheckCircle} \\
        prenatal-screens              & \textcolor{green}{\faCheckCircle} & \textcolor{green}{\faCheckCircle} \\
        problem-list              & \textcolor{darkgray}{\faTimesCircle} & \textcolor{green}{\faCheckCircle} \\
        procedure                     & \textcolor{green}{\faCheckCircle} & \textcolor{darkgray}{\faTimesCircle} \\
        procedure-history                 & \textcolor{darkgray}{\faTimesCircle} & \textcolor{green}{\faCheckCircle} \\
        reason                        & \textcolor{green}{\faCheckCircle} & \textcolor{darkgray}{\faTimesCircle} \\
        review-of-systems             & \textcolor{green}{\faCheckCircle} & \textcolor{green}{\faCheckCircle} \\
        social-and-family-history     & \textcolor{green}{\faCheckCircle} & \textcolor{darkgray}{\faTimesCircle} \\
        social-history                & \textcolor{green}{\faCheckCircle} & \textcolor{green}{\faCheckCircle} \\
        technique                     & \textcolor{green}{\faCheckCircle} & \textcolor{darkgray}{\faTimesCircle} \\
        wet-read                      & \textcolor{green}{\faCheckCircle} & \textcolor{darkgray}{\faTimesCircle} \\
        \hline
    \end{tabular}}
    \caption{Section Headers in MedSecId vs. ONC(\textcolor{green}{\faCheckCircle} = Present, \textcolor{darkgray}{\faTimesCircle} = Absent).}
    \label{tab:table_1}
\end{table}

\section{Methodology}
\label{sec:methods}
We explore two approaches for clinical section segmentation: Supervised Learning and Zero-shot Learning via LLMs. In this section, we provide an overview of both approaches.
% ; Detailed implementation, training configurations, and computational resource usage are provided in~Appendix~\ref{sec:appendix}.

\begin{table}[t]
\centering
\footnotesize
\resizebox{\columnwidth}{!}{
\begin{tabular}{lrr}
\toprule
             \textbf{Section Header} &  \textbf{Total Spans} &  \textbf{Overall \%} \\
\midrule
      social-history &          120 &       7.92 \\
 current-medications &          114 &       7.52 \\
           allergies &          114 &       7.52 \\
physical-examination &          102 &       6.73 \\
      family-history &           100 &       6.60 \\
   history-of-present-illness &           96 &       6.33 \\
   impression-and-plan &          82 &       5.41 \\
 chief-complaint &          80 &       5.28 \\
           review-of-systems &          79 &       5.21 \\
problem-list &          79 &       5.21 \\
      pregnancy-history &           79 &       5.21 \\
   gestational-age &           77 &       5.08 \\
   procedure-history &           67 &       4.42 \\
   past-medical-history &           62 &       4.09 \\
   labs &           52 &       3.43 \\
   past-surgical-history &           49 &       3.23 \\
   obstetrical-history &           47 &       3.10 \\
   gynecological-history &           46 &       3.03 \\
 assessment-and-plan &           20 &       1.32 \\
 critical-care-attending-addendum &           12 &       0.79 \\
 labs-imaging &           11 &       0.73 \\
 prenatal-history &           11 &       0.73 \\
 imaging &           10 &       0.66 \\
 obstetrical-and-gynecological-history &           2 &       0.13 \\
 plan &           1 &       0.07 \\
 prenatal-screens &           1 &       0.07 \\
 consent &           1 &       0.07 \\
 history-of-present-pregnancy &           1 &       0.07 \\
 prenatal-care &           1 &       0.07 \\
 
\bottomrule
\end{tabular}}
\caption{Frequency distribution of section headers in the ONC dataset  (excluding <none>).}
\label{tab:table_2}
\end{table}

\subsection{Supervised Learning Approach}
We first develop a supervised approach to clinical section segmentation using pre-trained transformer-based models, widely used in text classification and sequence labeling tasks \citep{vaswani2017attention, devlin2019bert}. While existing systems such as \citet{landes2022new} incorporate BiLSTMs and other architectures to achieve stronger performance, we focus on transformer-based baselines, reflecting prevailing supervised practice and offering a fair comparison for zero-shot LLMs.
We fine-tune the models using two architectures:

\begin{enumerate}
    \item \textbf {Transformer-based Classification: } Each line (i.e., a newline-separated sentence span extracted from the clinical note) is treated as an independent input and classified according to predefined section headers.
    
    \item \textbf {Transformer + CRF: } A Conditional Random Field (CRF) layer is added on top of the transformer to model label dependencies between consecutive lines, framing the task as sequence labeling.
\end{enumerate}

\subsubsection{Transformer-based Section Segmentation}
\label{sec:method1}
We approach section segmentation as a 51-way classification task (including the label \textit{"<none>"}) using an IO-like encoding scheme: lines within labeled sections are tagged as \textit{"I\_section\_name"}, while lines outside any labeled section are tagged as \textit{"<none>"} \citep{landes2022new}. Throughout this work, we use the terms \textit{"line"} and \textit{"sentence"} interchangeably, as each unit in our dataset corresponds to a single textual span separated by newlines in clinical notes. We experiment with BERT-base, a widely used transformer model pre-trained on general-domain English corpora \citep{devlin2019bert}, and three models trained on biomedical text. BioBERT \citep{lee2020biobert} extends BERT via further pretraining on PubMed abstracts and PubMed Central (PMC) articles. BiomedBERT (formerly PubMedBERT) \citep{gu2021domain} is trained from scratch exclusively on PubMed abstracts, making it a medical domain-specialized language model. GatorTron-base \citep{yang2022large} is trained on a diverse corpus comprising de-identified clinical notes from a university hospital, PubMed articles, and Wikipedia, totaling 90 billion words. We exclude models primarily trained on MIMIC-III (e.g., BioClinicalBERT \citep{alsentzer2019publicly}) to avoid evaluation bias.

\paragraph{Line-Level Representation.} 

We represent clinical notes as sequences of independent lines (rather than full-length notes) to comply with transformer token limits and reduce computational overhead. Each line is treated as a separate example, capturing local context without modeling sequential dependencies, resulting in a dataset of individual line-label pairs. Because the model does not leverage inter-line context, we also perform train-test splitting at the line level, consistent with the model's independence assumption and eliminating the need to preserve note boundaries. This process yields 175,703 lines from 2,002 clinical notes, with 80\% (140,140 lines) used for training and 20\% (35,563 lines) for evaluation. While this setup ignores document-level structure, it provides a fair supervised baseline for comparison with zero-shot LLMs, which do not explicitly model label transitions or structured dependencies across lines despite having access to the full note.

\paragraph{Token Length Analysis.} 
Before tokenization, we analyzed the distribution of token lengths per line. Approximately 97\% of lines (across all models) contained fewer than 100 subword tokens. We therefore truncate each line to 100 tokens to optimize memory usage and format inputs using the standard HuggingFace \citep{wolf2020transformers} convention: \textit{input\_ids} and \textit{attention\_mask} for training. 
Training configurations, hyperparameters, and evaluation metrics are provided in Appendix~\ref{sec:appendix1}.

\subsubsection{Transformer + CRF-based Section Segmentation}
Unlike the line-level approach in Section~\ref{sec:method1}, we retain note-level structure to model sequential dependencies between lines. Each note is treated as a single training instance, allowing the CRF layer to learn label transitions (e.g., from \textit{"History of Present Illness"} to \textit{"Review of Systems"}).

\paragraph{Custom Collator and Data Preparation.}
To accommodate varying note lengths, we implement a custom collator for note-level batching:

\begin{itemize}
    \item \textbf{Dynamic Line Dimensions:} For each batch, let \textit{L} be the maximum number of lines across notes; each line is truncated or padded to a maximum token length \textit{S}.
    \item \textbf {Batch-Size Constraint:} To preserve note-level context and avoid inefficient padding across variable-length sequences, we set batch size to \textit{B = 1}, which simplifies training and reduces GPU memory usage while retaining the CRF's ability to model label transitions.
     \item \textbf {Final Tensor Shape:} Each note is arranged into a tensor of shape \textit{(B,L,S)}, preserving full note structure. This allows the CRF to model label transitions across all lines within a note.
\end{itemize}

\paragraph{Model Architecture.} 
We use the same transformer backbones in Section~\ref{sec:method1} (BERT-base, BioBERT, BiomedBERT and GatorTron-base) combined with \textit{torchcrf}, a CRF library for PyTorch \citep{paszke2019pytorch}. The Transformer + CRF architecture consists of the following steps:

\begin{enumerate}
    \item \textbf {Flatten Input: } We reshape \textit{(B,L,S)} to \textit{(B x L,S)} so each line can be processed independently by the transformer.
    \item \textbf {Contextual Embeddings: } We extract the \textit{[CLS]} representation for each line.
    \item \textbf {Logit Projection: } We apply a linear layer to project contextual embeddings into logits of shape \textit{(B x L, num\_labels}) for each section label where \textit{num\_labels = 51}.
     \item \textbf {CRF Reshaping: } We reshape logits back to \textit{(B, L, num\_labels}), so the CRF can model line-level transitions across the entire note.
     \item \textbf {Viterbi Decoding: } At evaluation, we apply Viterbi decoding \citep{viterbi1967error} to obtain the most likely label sequence for each note.
 \end{enumerate}

Training hyperparameters and evaluation details are provided in Appendix~\ref{sec:appendix2}.

\subsection {Zero-Shot Learning via LLMs}
We explore zero-shot learning for clinical section segmentation using pre-trained LLMs. Our primary goal is to evaluate whether instruction-tuned LLMs without domain-specific fine-tuning can accurately assign section labels by leveraging general contextual understanding.

% in body:
\begin{figure}[t]
\centering
\lstset{style=mypromptstyle}
\begin{lstlisting}

<|begin_of_text|><|start_header_id|>system<|end_header_id|>
You are a clinical assistant specializing in segmenting clinical notes.

<|eot_id|><|start_header_id|>user<|end_header_id|>
Your task is to assign section headers to each line of a clinical note. Most of the section headers will likely span multiple lines, so headers should be assigned sequentially and consistently.

Clinical Note:
{enumerated clinical note text}

Select the most appropriate section header 
for each line from the following options:
{string of 30 potential headers}

Return your answer as a list of section headers, one for each line, in the same order.  

Example Output:
Line 0: <none>
Line 1: imaging
Line 2: <none>
Line 3: chief-complaint
Line 4: history-of-present-illness
Line 5: history-of-present-illness
Line 6: history-of-present-illness
Line 7: history-of-present-illness
Line 8: history-of-present-illness
Line 9: history-of-present-illness
...

The output must contain **exactly the same 
number of lines** as the clinical note, i.e., number of lines SHOULD BE EQUAL TO
{number of note lines}

<|eot_id|><|start_header_id|>assistant<|end_header_id|>
Section Headers:
\end{lstlisting}
\caption{Zero-shot prompt snippet for Llama Instruct models. The candidate label set corresponds to the 30 section headers defined in the ONC dataset; MedSecId uses a larger schema with 51 headers.}
\label{fig:zero_shot_prompt}
\end{figure}

\paragraph{Model Selection.}
We selected four instruction-tuned, open-source LLMs for evaluation: Mistral-7B-Instruct-v0.3 \citep{jiang2023mistral}, Llama 3.1-8B-Instruct \citep{touvron2023llama}, Qwen-2.5-32B-Instruct \citep{yang2024qwen2}, and Llama 3.3-70B-Instruct \citep{touvron2023llama}. These models support extended context windows (32k-128k tokens), enabling full-note inference without truncation. Their varied sizes (7B-70B) allow us to assess how model scale affects performance in long-form clinical narratives.

Although our dataset is de-identified, real-world clinical documents often contain PHI. Closed-source models such as GPT-4 \citep{openai2023gpt4} and Gemini \citep{team2023gemini} can pose security and privacy risks, as they require sending user data to third-party servers and thus increase the likelihood of unauthorized access or misuse of sensitive information \citep{kim2025benchmarking}. Open-source models can be deployed on-premise, offering a more secure pathway for integrating LLMs into clinical workflows. This practical consideration further motivates our use of open-source models.

\paragraph{Prompt Engineering.}
We adopt an instruction-style prompt to assign section labels to each line in a clinical note, without any task-specific fine-tuning. All four models are chat-based and support system/user prompting. The Llama models use a unified template with explicit system and user roles (see Figure~\ref{fig:zero_shot_prompt}); for Mistral and Qwen, we adapt the prompt format to match their respective syntax conventions (e.g., \textit{[INST]} or \textit{<|im\_start|>}). We designate the model as a “clinical assistant specializing in segmenting clinical notes” and provide it with a list of valid section labels. Each line in the note is numbered (e.g., “Line1,” “Line2”) to ensure independent prediction while preserving sequence order. This structure allows the model to reference neighboring lines during inference, enabling implicit modeling of section transitions. To clarify output formatting, we include a single one-shot-style example (e.g., \textit{Line 0: <none>, Line 1: imaging}). This preserves a near-zero-shot setup, relying solely on the model's pretrained knowledge to infer appropriate section labels. 
See Appendix~\ref{sec:appendix3} for inference details.

\paragraph{Postprocessing.}
We parse model outputs using regular expressions to isolate predicted section headers (e.g., removing “Line 0:” prefixes). Predictions are evaluated against gold labels in the MedSecId and ONC datasets using precision, recall, F1, and \textit{hallucination rate}, defined as the percentage of lines assigned to non-existent section headers. In collaboration with the midwifery expert who assisted with annotations, we consolidated \textit{impression-and-plan} and \textit{plan} into the standardized label \textit{assessment-and-plan}, following clinical convention. Because both labels were merged into an existing category, this consolidation reduced the effective ONC label set used for evaluation from 30 to 28 section headers. The resulting \textit{assessment-and-plan} label aligns with terminology adopted in prior clinical section segmentation work \citep{denny2009evaluation, landes2022new}, supporting its use as a canonical form for evaluation.

\section{Experiments}
\label{sec:experiments}

\subsection{Evaluation and Experimental Setup}
% We evaluate the performance of our supervised models and zero-shot LLMs on two datasets: MedSecId and ONC. Since the supervised models were trained on MedSecId, we exclude the training portion (80\%, or 1,601 notes) to avoid evaluation bias. From the remaining 401 notes, we further exclude those with more than 100 lines to maintain a tractable sequence length for evaluation, resulting in a final subset of 251 notes comprising 11,528 lines. For ONC, we used all 100 notes (5,352 lines).

We evaluate the performance of our supervised models and zero-shot LLMs on two datasets: MedSecId and ONC. For MedSecId, we evaluate on the test portion of the dataset (20\%, or 401 notes). To maintain a tractable sequence length for evaluation, we exclude notes with more than 100 lines, resulting in a final subset of 251 notes comprising 11,528 lines. 
For ONC, we use all 100 notes (5,352 lines).

\subsection{Hallucinations in Zero-Shot LLMs}

Despite receiving clear instructions, all four zero-shot models—Mistral-7B-Instruct-v0.3, Qwen-2.5-32B-Instruct, Llama 3.1-8B-Instruct, and Llama 3.3-70B-Instruct—exhibited hallucinations during inference by generating section headers not present in the ground truth. We define hallucination in this context as the assignment of a section header that does not appear in the predefined list of valid labels. For example, Mistral frequently labeled \textit{substance-abuse} as a section, although it should be  subsumed under the broader \textit{social history}. Such mislabeling risks fragmenting semantically related content, potentially compromising clinical workflows.

As shown in Table~\ref{tab:hallucination}, hallucination rates varied across models, with Mistral producing the highest rates on both datasets (22.21\% for MedSecId; 17.64\% for ONC), followed by Qwen, Llama 3.1-8B and Llama 3.3-70B. Interestingly, this ranking diverges from those reported in general-domain hallucination benchmarks (e.g., \citet{HughesBae2023}), underscoring the importance of evaluating model reliability within the specific context of clinical tasks. These findings suggest that hallucination behavior is highly sensitive to domain, task formulation, and prompting strategy, and cannot be reliably extrapolated from general-purpose evaluations. Further research is needed to address the factual consistency of LLM outputs in the healthcare domain \citep{nori2023capabilities}. 
To better characterize model behavior, Table~\ref{tab:hallucinated_headers} lists the five most frequently hallucinated section headers for each model on the ONC dataset.

\begin{table}[t]
    \centering
    \footnotesize % Reduce font size for better fit
    \renewcommand{\arraystretch}{1.2} % Adjust row height for readability
    \setlength{\tabcolsep}{3pt} % Reduce column spacing to fit within half-page

    \begin{tabular}{|c|c|c|c|}
        \hline
        \textbf{Model} & \textbf{HL} & \textbf{H\%} & \textbf{HS} \\ 
        \hline
        \multicolumn{4}{|c|}{\textbf{MedSecId}} \\
        \hline
        Mistral-7B-Instruct-v0.3 & 2,560 & 22.21\% & 433 \\
        Llama 3.1-8B Instruct & 452 & 3.92\% & 89 \\
        Qwen-2.5-32B-Instruct & 497 & 4.31\% & 54 \\
        Llama 3.3-70B Instruct & 404 & 3.50\% & 57 \\
        \hline
        \multicolumn{4}{|c|}{\textbf{ONC}} \\
        \hline
        Mistral-7B-Instruct-v0.3 & 944 & 17.64\% & 136 \\
        Llama 3.1-8B Instruct & 115 & 2.15\% & 19 \\
        Qwen-2.5-32B-Instruct & 177 & 3.31\% & 23 \\
        Llama 3.3-70B Instruct & 5 & 0.09\% & 4 \\
        
        \hline
    \end{tabular}
    \caption{Hallucination Analysis on MedSecId and ONC. \textit{HL} = number of hallucinated lines; \textit{H\%} = hallucination rate; \textit{HS} = number of hallucinated sections types not in the original label set.}
    \label{tab:hallucination}
\end{table}

\begin{table}[t]
\centering
\footnotesize
\renewcommand{\arraystretch}{1.3}
\begin{tabular}{|>{\centering\arraybackslash}p{0.25\linewidth}|>{\arraybackslash}p{0.65\linewidth}|}
\hline
\textbf{Model} & \textbf{Top 5 Hallucinated Sections} \\
\hline
Mistral-7B-Instruct-v0.3 & \textit{substance-abuse}, \textit{neurologic}, \textit{psychiatric}, \textit{psychosocial-history}, \textit{integumentary} \\
\hline
Llama 3.1-8B-Instruct & \textit{review / management}, \textit{review-and-management}, \textit{health maintenance}, \textit{psychosocial-history}, \textit{obstetrical-examination}  \\
\hline
Qwen-2.5-32B-Instruct & \textit{basic-information}, \textit{substance-abuse}, \textit{psychosocial-history}, \textit{obstetric-exam}, \textit{postoperative-information} \\
\hline
Llama 3.3-70B-Instruct & \textit{health-maintenance}, \textit{risk-factors}, \textit{psychosocial-history}, \textit{comments} \\
\hline
\end{tabular}
\caption{Top 5 most frequently hallucinated section headers generated by each model. (Llama 3.3-70B-Instruct produced only four hallucinated headers in total.)}
\label{tab:hallucinated_headers}
\end{table}

\paragraph{Mitigating Hallucinations.}

To mitigate hallucinations, we implemented a postprocessing correction step using GPT-4o \citep{openai2023gpt4}. For each hallucinated section header (i.e., one not present in the set of valid labels), we prompted GPT-4o to map it to the most semantically appropriate label from the valid list. Because this involved only generic section names (e.g., \textit{labs} or \textit{social-history}) and no patient-level content, we could safely use an API-based model without violating privacy constraints. We selected it over embedding-based heuristics (e.g., Sentence-BERT cosine similarity \citep{reimers-2019-sentence-bert}) due to its superior contextual reasoning, particularly for ambiguous or sparsely descriptive headers. 

While some edge cases remain challenging, this procedure substantially reduced the number of non-standard predictions and improved alignment with the target schema. Importantly, the correction accuracy may underestimate true semantic alignment: in some cases, hallucinated headers (e.g., \textit{ultrasound}) may be semantically closer to a different valid label (e.g., \textit{imaging}) than to the gold-standard label used for evaluation (e.g., \textit{review-of-systems}). In such cases, lower correction scores may reflect initial label misalignment rather than a failure of the mapping strategy. 
% We report the post-correction mapping results in Table~\ref{tab:hallucination_correction} (Appendix). The prompt used for GPT-4o hallucination correction is provided in Listing~\ref{fig:zero_shot_prompt_2} (Appendix).

\subsection{LLM-Based Error Analysis for Section Predictions}
After correcting hallucinations, we analyzed remaining section labeling errors through an automated evaluation of outputs from the best-performing model, Llama 3.3-70B-Instruct. To scale this process, we employed the same model in an LLM-based classification framework to assign errors to one of four categories: (1) \textit{Omission}, where the model incorrectly predicted \textit{<none>} for a span that should have received a valid label; (2) \textit{Label confusion}, where the model (under explicit prompt constraints) determined the predicted label was clearly incorrect relative to the gold label; (3) \textit{Valid local interpretation}, where the model judged the predicted label as semantically justifiable based on the local text span; and (4) \textit{Other}, capturing ambiguous or uncategorizable cases. The classification prompt handled categories 2–4, while \textit{Omission} was identified separately using rule-based logic. The full prompt, error distribution summary, and representative examples are provided in Appendix~\ref{appendix4} (Figure~\ref{fig:zero_shot_prompt2}, Figure~\ref{fig:qual_error_pie}, and Table~\ref{tab:error_examples}). 
% For brevity, we omit the prompt, which instructed the model to compare gold and predicted labels for the given text span and decide the error type.
% The prompt is shown in Figure~\ref{fig:zero_shot_prompt2} (Appendix). Figure~\ref{fig:qual_error_pie} summarizes error type distributions, and Table~\ref{tab:error_examples} provides representative examples; both appear in the Appendix.

\begin{table}[t]
\centering
\large
\renewcommand{\arraystretch}{1.3}
\setlength{\tabcolsep}{3pt}       % tighten columns a bit
\resizebox{\columnwidth}{!}{%
\begin{tabular}{|c|c|c|c|c|c|c|}
\hline
\textbf{Model} & \textbf{MP} & \textbf{MR} & \textbf{MF1} & \textbf{wP} & \textbf{wR} & \textbf{wF1} \\
\hline
\multicolumn{7}{|c|}{\textbf{Supervised Models}} \\
\hline
BERT\textsubscript{base} & 0.71 & 0.67 & 0.68 & 0.78 & 0.78 & 0.77 \\
BioBERT & 0.72 & 0.68 & 0.68 & 0.78 & 0.78 & 0.77 \\
BiomedBERT & 0.72 & 0.69 & 0.68 & 0.79 & 0.79 & 0.78 \\
GatorTron\textsubscript{base} & 0.73 & 0.69 & \textbf{0.69} & 0.80 & \textbf{0.80} & 0.78 \\
BERT\textsubscript{base}+CRF & 0.72 & 0.69 & 0.68 & 0.79 & 0.77 & 0.77 \\
BioBERT+CRF & 0.74 & 0.69 & 0.68 & 0.79 & 0.77 & 0.76 \\
BiomedBERT+CRF & \textbf{0.75} & \textbf{0.70} & \textbf{0.69} & 0.79 & 0.79 & 0.78 \\
GatorTron\textsubscript{base}+CRF & 0.74 & 0.65 & 0.67 & \textbf{0.81} & \textbf{0.80} & \textbf{0.79} \\
\hline
\multicolumn{7}{|c|}{\textbf{Zero-Shot Models (Results with Hallucinations)}} \\
\hline
Mistral-7B-Instruct\textsubscript{raw} & 0.03 & 0.02 & 0.02 & 0.54 & 0.17 & 0.22 \\
Llama 3.1-8B-Instruct\textsubscript{raw} & 0.17 & 0.19 & 0.14 & 0.70 & 0.48 & 0.52 \\
Qwen-2.5-32B-Instruct\textsubscript{raw} & 0.25 & 0.23 & 0.21 & 0.68 & 0.45 & 0.49 \\
Llama 3.3-70B-Instruct\textsubscript{raw} & 0.23 & 0.29 & 0.23 & 0.76 & 0.61 & 0.64 \\
\hline
\multicolumn{7}{|c|}{\textbf{Zero-Shot Models (Results after Mitigating Hallucinations)}} \\
\hline
Mistral-7B-Instruct\textsubscript{corrected} & 0.21 & 0.19 & 0.16 & 0.41 & 0.20 & 0.23 \\
Llama 3.1-8B-Instruct\textsubscript{corrected} & 0.46 & 0.54 & 0.39 & 0.70 & 0.49 & 0.52 \\
Qwen-2.5-32B-Instruct\textsubscript{corrected} & 0.47 & 0.48 & 0.41 & 0.61 & 0.46 & 0.49 \\
Llama 3.3-70B-Instruct\textsubscript{corrected} & 0.47 & 0.61 & 0.48 & 0.73 & 0.62 & 0.64 \\
\hline
\end{tabular}%
}
\caption{Performance metrics on MedSecId: MP = macro precision, MR = macro recall, MF1 = macro F1; wP = weighted precision, wR = weighted recall, wF1 = weighted F1.}
\label{tab:results_1}
\end{table}

\begin{table}[t]
\centering
\large
\renewcommand{\arraystretch}{1.3}
\setlength{\tabcolsep}{3pt}       % tighten columns a bit
\resizebox{\columnwidth}{!}{%
\begin{tabular}{|c|c|c|c|c|c|c|}
\hline
\textbf{Model} & \textbf{MP} & \textbf{MR} & \textbf{MF1} & \textbf{wP} & \textbf{wR} & \textbf{wF1} \\
\hline
\multicolumn{7}{|c|}{\textbf{Supervised Models}} \\
\hline
BERT\textsubscript{base} & 0.66 & 0.37 & 0.39 & 0.76 & 0.43 & 0.47 \\
BioBERT & 0.54 & 0.39 & 0.39 & 0.75 & 0.45 & 0.48 \\
BiomedBERT & 0.61 & 0.39 & 0.40 & 0.76 & 0.46 & 0.49 \\
GatorTron\textsubscript{base} & \textbf{0.73} & 0.48 & 0.49 & 0.85 & 0.58 & 0.61 \\
BERT\textsubscript{base}+CRF & 0.68 & 0.49 & 0.47 & 0.80 & 0.61 & 0.62 \\
BioBERT+CRF & 0.55 & 0.45 & 0.43 & 0.74 & 0.59 & 0.57 \\
BiomedBERT+CRF & 0.56 & 0.51 & 0.50 & 0.76 & 0.65 & 0.66 \\
GatorTron\textsubscript{base}+CRF & 0.65 & 0.51 & 0.49 & 0.79 & 0.65 & 0.65 \\
\hline
\multicolumn{7}{|c|}{\textbf{Zero-Shot Models (Results with Hallucinations)}} \\
\hline
Mistral-7B-Instruct\textsubscript{raw} & 0.05 & 0.04 & 0.04 & 0.72 & 0.45 & 0.52 \\
Llama 3.1-8B-Instruct\textsubscript{raw} & 0.35 & 0.33 & 0.32 & 0.84 & 0.70 & 0.74 \\
Qwen-2.5-32B-Instruct\textsubscript{raw} & 0.34 & 0.39 & 0.34 & 0.88 & 0.79 & 0.83 \\
Llama 3.3-70B-Instruct\textsubscript{raw} & 0.61 & 0.59 & 0.58 & \textbf{0.90} & \textbf{0.85} & \textbf{0.86} \\
\hline
\multicolumn{7}{|c|}{\textbf{Zero-Shot Models (Results after Mitigating Hallucinations)}} \\
\hline
Mistral-7B-Instruct\textsubscript{corrected} & 0.38 & 0.45 & 0.37 & 0.56 & 0.47 & 0.49 \\
Llama 3.1-8B-Instruct\textsubscript{corrected} & 0.58 & 0.56 & 0.54 & 0.83 & 0.71 & 0.74 \\
Qwen-2.5-32B-Instruct\textsubscript{corrected} & 0.61 & \textbf{0.71} & 0.61 & 0.88 & 0.82 & 0.84 \\
Llama 3.3-70B-Instruct\textsubscript{corrected} & 0.70 & 0.67 & \textbf{0.67} & \textbf{0.90} & \textbf{0.85} & \textbf{0.86} \\
\hline
\end{tabular}%
}
\caption{Performance metrics on ONC: MP = macro precision, MR = macro recall, MF1 = macro F1; wP = weighted precision, wR = weighted recall, wF1 = weighted F1.}
\label{tab:results_2}
\end{table}

\subsection{Results and Discussion}

Tables~\ref{tab:results_1} and~\ref{tab:results_2} present the performance of all models on the MedSecId and ONC datasets, respectively. For zero-shot LLMs, we report both raw and corrected results to highlight the impact of hallucination mitigation. Notably, post-correction macro F1 scores increase by 9\% to 33\%, confirming that hallucinations are a major source of error in zero-shot predictions.

As expected, supervised models outperform zero-shot LLMs on MedSecId due to their direct training on that dataset. Among the supervised models, performance is largely comparable across transformer variants. However, the addition of a CRF layer yields modest but non-negligible gains for some models. Specifically, macro F1 scores for BERT-based models improve by 4\% to 10\% with CRF integration, suggesting that modeling inter-line dependencies offers measurable benefits. In contrast, GatorTron shows no improvement with a CRF layer, indicating that larger models may already encode sufficient contextual information for accurate line-level predictions. Meanwhile, zero-shot LLMs display large discrepancies between macro and weighted F1 scores due to macro F1's sensitivity to hallucinated labels. Once hallucinated headers are corrected, spurious labels are mapped to valid alternatives, leading to substantial improvements in macro-averaged performance scores.

While supervised models maintain a strong lead over zero-shot LLMs on MedSecId, they struggle to generalize to the new ONC dataset. This suggests that models trained on large public corpora, such as MIMIC, may not transfer effectively to narrower clinical subdomains. Although GatorTron-base initially outperforms other supervised models, the addition of a CRF layer allows others to close or surpass the gap. Notably, BioMedBERT+CRF outperforms GatorTron+CRF by approximately 1\% in macro and weighted F1 on the ONC dataset.

Interestingly, zero-shot LLMs perform relatively better on ONC, partly due to the smaller label space (28 versus 51 labels). To quantify the robustness of model performance across notes, we computed 95\% confidence intervals for per-note macro and weighted F1 (Table~\ref{tab:ci_results}), which confirmed the stability of reported trends. Llama 3.3-70B-Instruct achieves the highest overall performance, outperforming all supervised baselines. To assess the consistency of this advantage, we conducted Wilcoxon signed-rank tests on per-note macro F1. Even in its hallucinated form, Llama 3.3-70B-Instruct significantly outperforms the strongest supervised model (\(p < 4.88 \times 10^{-17} \)), with further gains after correction (\( p < 3.75 \times 10^{-17} \)). These results suggest that the LLM's advantage reflects robust generalization, not merely post-hoc label correction. While Llama 3.3-70B-Instruct slightly outperforms Qwen-2.5-32B-Instruct on average, the difference is not statistically significant (\( p \approx 0.11 \)), indicating comparable performance between the strongest zero-shot models. 

Overall, these findings highlight the flexibility of zero-shot LLMs in adapting to novel domains without requiring additional annotation or fine-tuning. While supervised transformer models remain state-of-the-art for in-domain tasks, instruction-tuned LLMs offer a statistically robust and scalable alternative for clinical NLP in underexplored subdomains, especially when paired with simple hallucination correction.

% \begin{table}[t]
% \centering
% \footnotesize
% \renewcommand{\arraystretch}{1.3}
% \begin{tabular}{|>{\centering\arraybackslash}p{0.25\linewidth}|>{\arraybackslash}p{0.65\linewidth}|}
% \hline
% \textbf{Model} & \textbf{Top 5 Hallucinated Sections} \\
% \hline
% Mistral-7B-Instruct-v0.3 & \textit{substance-abuse}, \textit{neurologic}, \textit{psychiatric}, \textit{psychosocial-history}, \textit{integumentary} \\
% \hline
% Llama 3.1-8B-Instruct & \textit{review / management}, \textit{review-and-management}, \textit{health maintenance}, \textit{psychosocial-history}, \textit{obstetrical-examination}  \\
% \hline
% Qwen-2.5-32B-Instruct & \textit{basic-information}, \textit{substance-abuse}, \textit{psychosocial-history}, \textit{obstetric-exam}, \textit{postoperative-information} \\
% \hline
% Llama 3.3-70B-Instruct & \textit{health-maintenance}, \textit{risk-factors}, \textit{psychosocial-history}, \textit{comments} \\
% \hline
% \end{tabular}
% \caption{Top 5 most frequently hallucinated section headers generated by each model. (Llama 3.3-70B-Instruct produced only four hallucinated headers in total.)}
% \label{tab:hallucinated_headers}
% \end{table}

\begin{table}[t]
\centering
\footnotesize
\renewcommand{\arraystretch}{1.3}
\begin{tabular}{|>{\centering\arraybackslash}p{0.35\linewidth}|>{\centering\arraybackslash}p{0.23\linewidth}|>{\centering\arraybackslash}p{0.23\linewidth}|}
\hline
\textbf{Model} & \textbf{Macro F1 (±95\% CI)} & \textbf{Weighted F1 (±95\% CI)} \\
\hline
Llama 3.3-70B-Instruct\textsubscript{corrected} & 0.800 ± 0.024 & 0.851 ± 0.024 \\
\hline
Qwen-2.5-32B-Instruct\textsubscript{corrected}  & 0.764 ± 0.032 & 0.818 ± 0.038 \\
\hline
BioMedBERT+CRF                                   & 0.604 ± 0.026 & 0.646 ± 0.028 \\
\hline
\end{tabular}
\caption{Comparison of per-note macro and weighted F1 scores ± 95\% bootstrap confidence intervals across 100 obstetric notes for the top two zero-shot LLMs and the best-performing supervised model (BioMedBERT+CRF).}
\label{tab:ci_results}
\end{table}

\section{Conclusions and Future Work}
\label{sec:conclusion}
In this work, we addressed clinical section segmentation in a specialized domain by introducing the \textit{Obstetrics Notes Collection (ONC)}. We evaluated supervised and zero‐shot LLM segmentation approaches on this dataset and a widely used public corpus. We found that while supervised models perform well in-domain, they struggle to generalize to unfamiliar clinical subdomains, whereas zero-shot LLMs show greater adaptability when domain-specific fine-tuning is unavailable.

Despite these advances, several challenges remain. Our dataset's limited size may not capture the full variability of obstetrics documentation, and although zero‐shot LLMs reduce reliance on labeled data, they remain prone to domain-inconsistent predictions such as hallucinated section headers and omissions of clinically important spans. These issues are concerning, as mislabeling or omitting clinically important spans can reduce reliability and interpretability.

Future work includes expanding the dataset to cover a wider range of conditions, procedures, and patient profiles, improving clinical diversity. We will also explore further LLM adaptation strategies, such as few‐shot learning and parameter‐efficient fine‐tuning (PEFT), to more effectively tailor models to specialized domains while retaining computational efficiency \citep{han2024parameter}. Finally, integrating domain knowledge bases or medical ontologies may enhance performance and interpretability by guiding segmentation and label assignment. These efforts aim to support the development of robust, domain-aware clinical NLP systems.

% \section{Limitations}
% \label{sec:limit}
% %Due to time and resource constraints, 
% ONC currently includes 100 H\&P narratives (50 from VBAC patients and 50 from RCS patients), randomly selected from a larger pool. While this subset provides an initial snapshot of obstetrics‐focused documentation, it may not capture the full variability of patients in this domain. For section annotation, we adopted a set of obstetrics-specific headers developed in collaboration with a certified midwifery expert. While these labels offer improved clinical relevance over general-purpose schemas such as MedSecId \citep{landes2022new}, they may introduce subjectivity, as other experts might define or group sections differently. This lack of standardization may limit comparability across datasets or models. Finally, our error categorization step relied on the same LLM that produced the outputs, which could in principle introduce model-specific bias; however, the coarse categorization scheme and rule-based handling of omissions mitigate the risk. Future work should explore building consensus-driven or ontology-aligned section schemas tailored to obstetrics, as well as expanding dataset coverage to better reflect diverse clinical structures and documentation styles. 

\section{Ethics Statement and Limitations}
\label{sec:limit}

\subsection{Ethical Considerations}
This study uses de-identified clinical narratives derived from electronic health records. As described in Sec.~\ref{sec:data}, all notes were processed within a HIPAA-compliant secure research environment and underwent automated and manual de-identification prior to analysis. The study was conducted under approval from the University of Illinois Chicago Institutional Review Board (IRB \#2022-0139). The publicly shared Obstetrics Notes Collection (ONC) dataset was additionally reviewed by the University of Illinois Chicago HIPAA Privacy Office to ensure compliance with data protection requirements.

\subsection{Limitations}
ONC currently includes 100 H\&P narratives (50 from VBAC patients and 50 from RCS patients), randomly selected from a larger pool. While this subset provides an initial snapshot of obstetrics‐focused documentation, it may not capture the full variability of patients in this domain. For section annotation, we adopted a set of obstetrics-specific headers developed in collaboration with a certified midwifery expert. While these labels offer improved clinical relevance over general-purpose schemas such as MedSecId \citep{landes2022new}, they may introduce subjectivity, as other experts might define or group sections differently. This lack of standardization may limit comparability across datasets or models. Finally, our error categorization step relied on the same LLM that produced the outputs, which could in principle introduce model-specific bias; however, the coarse categorization scheme and rule-based handling of omissions mitigate the risk. Future work should explore building consensus-driven or ontology-aligned section schemas tailored to obstetrics, as well as expanding dataset coverage to better reflect diverse clinical structures and documentation styles.

\section*{Acknowledgments}
\label{sec:acknowledge}
We thank Joanna Tess and Subhash Kumar Kolar from the University of Illinois Center for Clinical and Translational Science (CCTS) for their assistance with extracting the obstetric clinical notes used as the source corpus from which the ONC dataset was curated.

\clearpage % Forcing a new page for references
%\section{Bibliographical References}
\section{References}
\label{sec:reference}
\bibliographystyle{lrec2026-natbib}
\bibliography{karacan-lrec2026}

\clearpage % Forcing a new page for appendix
%\FloatBarrier
\appendix

\section{Appendix}
\label{sec:appendix}

\subsection{Transformer-Based Section Segmentation—Training and Evaluation }
\label{sec:appendix1}

\paragraph{Training Configuration} 
We use the Trainer class from HuggingFace \citep{wolf2020transformers} with the following hyperparameters (tuned within our GPU/memory constraints):

\begin{itemize} [noitemsep]
    \item \textbf{Learning rate:} 2e-5
    \item \textbf {Epochs:} 5
    \item \textbf {Batch size:} 32
     \item \textbf {Mixed precision:} Training is accelerated with bf16 precision
     \item \textbf {Max token length:} 100
     \item \textbf{Warmup steps:} 500
    \item \textbf {Weight decay:} 0.05
\end{itemize}

\paragraph{Evaluation Metrics} 
We compute standard classification metrics—accuracy, precision, recall, F1—along with macro-F1 (class-agnostic) and weighted-F1 (weighing classes by frequency) to assess how class imbalance affects performance.

\subsection{Transformer + CRF-Based Section Segmentation—Training and Evaluation }
\label{sec:appendix2}

\paragraph{Training Details} The training details for our experiments are as follows:

\begin{itemize} [noitemsep]
    \item \textbf{Learning rate:} 2e-5
     \item \textbf{Epoch:} 5
     \item \textbf {Batch size:} \textit{B = 1}
    \item \textbf {Mixed precision:} Training is accelerated with bf16 precision
    \item \textbf{Optimizer:} AdamW \citep{loshchilov2019decoupled} (updates Transformer + CRF parameters)
    \item \textbf {Max token length:} 100 for BERT-base, BioBERT, BiomedBERT; 64 for GatorTron (due to higher memory consumption)
\end{itemize}

\paragraph{Evaluation Metrics} 
As in Section~\ref{sec:appendix1}, we compute precision, recall, macro-F1, and weighted-F1 to evaluate note-level segmentation performance.

\subsection{Zero-Shot Learning via LLMs—Inference Details }
\label{sec:appendix3}

\paragraph{Inference Details}
To generate section labels, we perform a forward pass in inference-only mode with the following parameters:

\begin{itemize} [noitemsep]
    \item \textbf{temperature = 0.0}: Forces greedy decoding, prioritizing the most probable token at each step for consistent and deterministic output.
     \item \textbf{do\_sample = False}: Disables random sampling, ensuring reproducible outputs for identical prompts.
     \item \textbf {num\_beams = 1}: Avoids complex beam search, reducing computational overhead.
    \item \textbf {pad\_token\_id = tokenizer.eos\_token\_id}: Uses the end-of-sequence token for padding, preventing extraneous tokens in the output.
\end{itemize}

By combining these settings, our inference procedure remains deterministic and focused, yielding consistent line-by-line label predictions for each clinical note.

\subsection{Qualitative Error Analysis}
\label{appendix4}

% \paragraph{Distribution of Error Types (Llama 3.3-70B-Instruct Model)}
\begin{figure}[H]
\centering
\begin{tikzpicture}
\pie[
    radius=1.8, % smaller radius for single-column width
    text=legend,
    %text style={font=\tiny}, % smaller font for labels
    %text style={font=\footnotesize},
    font=\footnotesize,
    color={
        blue!70,
        orange!85,
        green!60!black,
    }
]{
    4/Omission,
    64/Label Confusion,
    32/VL Interpretation
}
\end{tikzpicture}
\caption{Proportional distribution of section labeling errors for Llama 3.3-70B-Instruct (Other = 0\%, 0 instances).}
\label{fig:qual_error_pie}
\end{figure}

% in body:
\begin{figure*}[t]
\centering
\lstset{style=mypromptstyle}
\begin{lstlisting}

<|begin_of_text|><|start_header_id|>system<|end_header_id|>
You are analyzing a prediction error in a clinical note section classification task. A sentence was assigned a gold-standard label. A language model attempted to predict this label.

<|eot_id|><|start_header_id|>user<|end_header_id|>
Your task is to assign section headers to each line of a clinical note. Most of the section headers will likely span multiple lines, so headers should be assigned sequentially and consistently.

You are given:
    - Gold label:{gold_label}
    - Predicted label: {predicted_label}
    - Text span: {span_text}

Your task is to decide what type of error this is.
    
Choose only one of the following categories:
1. **Label Confusion** The model predicted a valid but clearly different label from the gold.
2. **Valid Local Interpretation** The predicted label is different from gold, but makes semantic sense given the span alone.
3. **Other** This case is ambiguous or doesn't fit the above categories.
    
Respond exactly in the following format:
Label: <one of the 3 options above>  
Reason: <your brief explanation>

<|eot_id|><|start_header_id|>assistant<|end_header_id|>
Section Headers:
\end{lstlisting}
\caption{Zero-shot prompt snippet for Llama Instruct models}
\label{fig:zero_shot_prompt2}
\end{figure*}

\begin{table*}[t]
\centering
\footnotesize
\begin{tabular}{|>{\centering\arraybackslash}p{0.3\linewidth}|
                >{\centering\arraybackslash}p{0.2\linewidth}|
                >{\centering\arraybackslash}p{0.45\linewidth}|}
\hline
\textbf{Text Span} & \textbf{Error Type} & \textbf{Explanation} \\
\hline
Review / Management  Results review:   GBS (group B streptococcal): negative     Hepatitis B : negative Syphilis screen: NR x2 Rubella: Immune     HIV: NR      STDs: Neg      Blood type: O+ 1-hr GTT: 94      Genetic Screening Tests (First/Sequential/QUAD) : normal. & Label Confusion & 	The predicted label "assessment-and-plan" is a valid label, but it clearly differs from the gold label "labs-imaging", as the text span primarily discusses laboratory results, which aligns more closely with "labs-imaging". \\
\hline
In addition patient was incidentally found to have 4cm arrachnoid cyst of her left temporal fossa. & Valid Local Interpretation & The predicted label \textit{physical-examination} makes sense given the text span, which describes a medical finding, even though the gold label is \textit{assessment-and-plan}. The sentence could be part of a physical examination section, but in the context of the entire clinical note, it might be more appropriately classified under assessment and plan due to the incidental finding mentioned. \\
\hline
\end{tabular}
\caption{Representative examples of Llama 3.3-70B-Instruct prediction mismatches categorized by LLM, with associated reasoning.}
\label{tab:error_examples}
\end{table*}

\end{document}